\documentclass[
  onecolumn,
  natbib,
]{miri-tech-article}

\usepackage[utf8]{inputenc} 

\usepackage{enotez}
\setenotez{backref = true}
\usepackage[T1]{fontenc}
\usepackage{lmodern}
\usepackage{float}
\usepackage{amsfonts, amsmath}       
\usepackage{float}
\usepackage{graphicx} 
\usepackage{bold-extra} 
\usepackage{bm}
\usepackage{graphicx}
\usepackage{enumerate}
\usepackage{enumitem}
\usepackage{array}
\usepackage[noabbrev]{cleveref}
\usepackage[font=small,labelfont=bf]{caption}
\usepackage{color, colortbl}	
\definecolor{Gray}{gray}{0.9}
\usepackage[hang,flushmargin]{footmisc}
\usepackage{setspace}
\usepackage{hyperref}
 \hypersetup{
     colorlinks=true,
     linkcolor=blue,
     filecolor=blue,
     citecolor = blue,      
     urlcolor=cyan,
     }
     
\usepackage{xpatch}

\newcommand{\myitem}[2]{  
\begin{enumerate}[leftmargin=31pt,nosep,topsep={0pt},partopsep={0pt},before={\begin{minipage}[t]{\hsize}\raggedright}, after={\end{minipage}}]
        \item[#1] #2 
    \end{enumerate}%
    }
\newcommand{\myitemize}[1]{  
\begin{enumerate}[leftmargin=31pt,nosep,topsep={0pt},partopsep={0pt},before={\begin{minipage}[t]{\hsize}\raggedright}, after={\end{minipage}}]
        #1
    \end{enumerate}%
    }    

\usepackage{glossaries}
\makeglossaries
\newglossaryentry{autostigmergy}
{
    name=autostigmergy,
    description={Using the environment as an external memory system (see ‘Externalizing representations’}
}
\newglossaryentry{cognitosecurity}
{
    name=cognitosecurity,
    description={Cognitosecurity is the domain concerned with protecting thoughts from mind-reading adversaries.}
}
\newglossaryentry{cognito-defense}
{
    name=cognito-defense,
    description={When the interpreted agent hides its thoughts from an adversary by exploiting weaknesses in the interpretability methods.}
}
\newglossaryentry{cognito-hazard}
{
    name=cognito-hazard,
    description={An observation that is dangerous to the observer merely by being perceived.}
}
\newglossaryentry{cognito-offense}
{
    name=cognito-offense,
    description={The act of seeking access to an adversary’s thoughts.}
}
\newglossaryentry{exposed thought-surface}
{
    name=exposed thought-surface,
    description={The thoughts that are theoretically legible by a interpretability tool or human interpreter.}
}
\newglossaryentry{interpretability method}
{
    name=interpretability method,
    description={The process of translating representations in an AI to representations in human interpreters. Contains interpretability tools and human interpreters.}
}
\newglossaryentry{interpretability tools}
{
    name=interpretability tools,
    description={The machinery that reads the AI's representations and converts them into outputs that are readable to humans. Components of interpretability methods.}
}
\newglossaryentry{xenosemanticity}
{
    name=xenosemanticity,
    description={The defining property of ‘alien representations’, where the meaning of a representation sits in a strange place in our (human) ontology.}
}

\usepackage{listings} 
\usepackage{tcolorbox}
\usepackage{realboxes}
\usepackage[export]{adjustbox}
\definecolor{mygreen}{rgb}{0,0.6,0}
\definecolor{mygray}{rgb}{0.5,0.5,0.5}
\definecolor{mymauve}{rgb}{0.58,0,0.82}
\definecolor{mygray}{rgb}{0.8,0.8,0.8}
\usepackage{placeins}

\makeatletter
\xpretocmd\lstinline{\Colorbox{mygray!40}\bgroup\appto\lst@DeInit{\egroup}}{}{}
\makeatother

\title{Circumventing interpretability: How to defeat mind-readers}
\author{Lee Sharkey\thanks{I’m grateful to David Lindner, Evan R. Murphy, Alex Lintz, Sid Black, Kyle McDonnell, Laria Reynolds, Adam Shimi, and Daniel Braun whose comments greatly improved earlier drafts of this article. The article’s weaknesses are mine, but many of its strengths are due to their contributions. Additionally, this article benefited from the prior work of many authors, but especially: Evan Hubinger, Peter Barnett, Adam Shimi, Neel Nanda, Evan R. Murphy, Eliezer Yudkowsky, Chris Olah. I collected several of the potential circumvention methods from their work. Part of this work was carried out while at Conjecture. The original post on which this paper was based can be found \href{https://www.alignmentforum.org/posts/EhAbh2pQoAXkm9yor/circumventing-interpretability-how-to-defeat-mind-readers}{here}.}}
\renewcommand\footnotemark{}

\date{December 2022}

\begin{document}
\maketitle
\setcounter{footnote}{0} 

\begin{abstract}
    The increasing capabilities of artificial intelligence (AI) systems make it ever more important that we interpret their internals to ensure that their intentions are aligned with human values. Yet there is reason to believe that misaligned artificial intelligence will have a convergent instrumental incentive to make its thoughts difficult for us to interpret. In this article, I discuss many ways that a capable AI might circumvent scalable interpretability methods and suggest a framework for thinking about these potential future risks.
\end{abstract}

\clearpage
{
\singlespacing
\small
\setcounter{tocdepth}{3}
\tableofcontents
}

\clearpage

\section*{Introduction}
\addcontentsline{toc}{section}{Introduction}

Many different AI alignment proposals have highlighted the importance of being able to interpret our AI models \citep{hub2020overview11}. However, it has so far proven difficult to comprehensively understand the representations\endnote{Without getting bogged down in terminology, I’m talking about ‘representations’ in a functional sense: A representation is just a vector in neural statespace that does a particular thing in a particular context; it’s a representation \emph{for} whatever its function is. It doesn’t necessarily need to generalize; nearby directions don’t need to perform similar functions and it might fail to perform that function in slightly different contexts.} that neural networks use, which poses a barrier to ensuring their safety \citep{toymodels2022}. Interpreting an unaligned AI may introduce an additional challenge: The AI would have a convergent instrumental incentive \citep{bts2016convergent} \endnote{I was inspired to write this article after I saw Yudkowsky complain \citep{soares21} that current AI safety discourse failed to inhabit even a shallowly adversarial security mindset with regard to using transparency/interpretability to make AI safe. He said that you should \emph{``\dots read fantasy novels about demons and telepathy, if you want a better appreciation of the convergent incentives of agents facing mindreaders than the 'AI safety' field outside myself is currently giving you.''} Soeren Elverlin asked reddit \citep{soerenelverlin2021} for recommendations that fitted Yudkowsky’s description and got plenty of answers. While that quote was definitely my inspiration, I didn't think reading those books would be more valuable or sufficiently relevant to AI interpretability than a few days of hard thinking about the problem (and my guess is that Yudkowsky wasn’t suggesting those books would be better than directly thinking about the problem). After a few days' thinking, I found that I'd come up with all of the mind-reading circumvention methods from those books that redditors described. So this article at least includes most of the obvious circumvention methods.} to circumvent our interpretability methods. There exists some sporadic past discussion about how an AI might achieve this \citep[e.g.][]{nanda22,evhub19,peterbarnett21,soerenelverlin2021,peterbarnett21a,soares21,murphy22}. But to my knowledge the problem hasn’t been addressed systematically, at least not publicly.

If AI safety researchers continue to make large bets on interpretability, then the lack of attention on potential circumvention methods may eventually lead to problems. We might end up with a false sense of security if interpretability methods say \emph{“No dangerous thoughts here!”} when in fact the AI has simply found a way to think dangerous thoughts undetected. 

We can think of this problem as the focus of an interpretability research agenda that is related to but distinct from the well-known agenda that asks \emph{“Can we understand what today’s neural networks are doing? And how can we scale our methods for understanding them?”.} This is clearly an important agenda. But we should probably also try to anticipate future issues by building a longer term interpretability research agenda. This article helps to address that longer term agenda and asks \emph{“Assuming our short term research agenda at least partly succeeds, how might a capable AI circumvent our interpretability methods?”.} 

Here I present a definitely-incomplete list of potential interpretability circumvention methods. Some are original, though many are collected from other sources. Throughout, I assume we’re concerned with interpreting the representations of a capable (roughly human-level or above) deep learning-based AI. I also suggest a framework for thinking about these circumvention methods, which I hope will help to identify omitted circumvention methods before a capable AI does.

\subsection*{The framework: Interpretation as information flow}
\addcontentsline{toc}{subsection}{The framework: Interpretation as information flow}

\begin{figure}[!ht]
    \centering
    \includegraphics[width=0.85\linewidth]{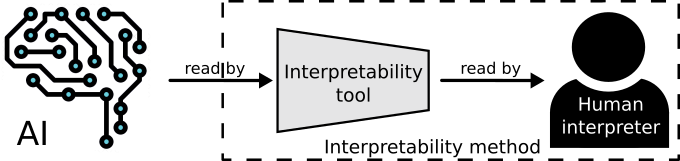}
    \caption{A simple model of information flow in interpretability.}
    \label{fig:Fig1}
\end{figure}

We can think of interpretation as information flow (Figure~\ref{fig:Fig1}). Here we’ll use the following basic model of information flow in interpretability:
\begin{itemize}
    \item We have an AI whose thoughts are being interpreted by an interpretability method.
    \item An \emph{interpretability method} is the process of translating an AI’s representations into whatever the corresponding representations are in humans.
    \item To do this, interpretability methods contain \emph{\gls{interpretability tools}}, which is the analysis machinery that reads the AI's representations and converts them into something legible to humans.
    \item \emph{Human interpreters} read the outputs of the \gls{interpretability tools} and attempt to understand them, thus converting them into human representations.
\end{itemize}

Throughout the article, I assume we’ve figured out ways to interpret our models at scale, which probably means that our \gls{interpretability tools} are automating a lot of the work. Nevertheless, all the potential circumvention methods I talk about apply to worlds where we still haven’t managed to scale interpretability, where the tools would be relatively less important.

\begin{figure}[!ht]
    \centering
    \includegraphics[width=0.85\linewidth]{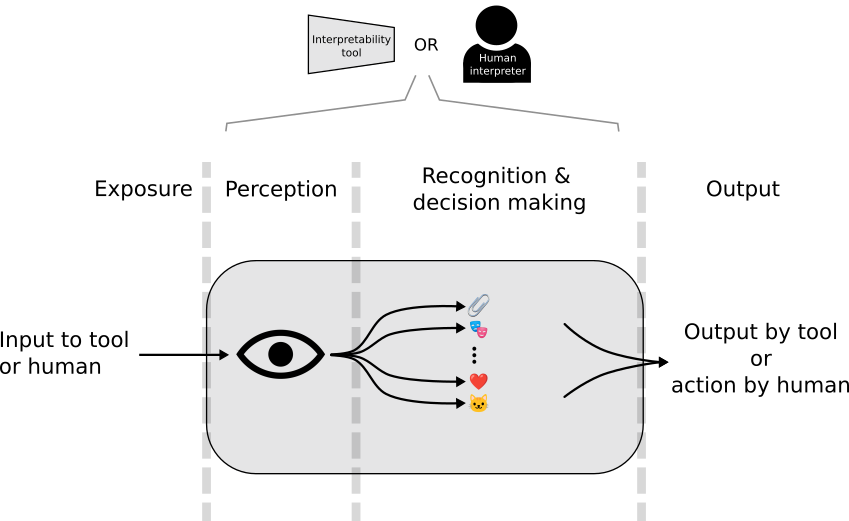}
    \caption{A more detailed model of information flow which is applicable to both the interpretability tool and to the human. Information flow begins with exposure of the information to the observer, then perception by the observer, then recognition \& and decision making, followed by output.}
    \label{fig:Fig2}
\end{figure}

An even more detailed model of information flow (Figure~\ref{fig:Fig2}) is useful for categorizing interpretability circumvention methods. Information flows within both the \textit{interpretability tool} and the \textit{human}. In both,

\begin{itemize} 
    \item What is to be interpreted must first be \emph{exposed} to the tool or the human.  
    \item What is exposed must then be \emph{perceived} or detected.
    \item What is perceived is then further analysed so that it can be \emph{recognized} or understood, which then determines an appropriate output by the tool or action by the human.
\end{itemize}

\subsection*{How I categorize interpretability circumvention methods}
\addcontentsline{toc}{subsection}{How I categorize interpretability circumvention methods}

The ways to categorize interpretability circumvention methods that I’ve found most useful (and the way this article is structured) are:

\subsubsection*{A. Exploiting weaknesses of the tool vs. weaknesses of the human interpreter}
\addcontentsline{toc}{subsubsection}{A. Exploiting weaknesses of the tool vs. weaknesses of the human interpreter}

Interpretability circumvention methods can either circumvent the \emph{interpretability tool} or the \emph{human interpreter}. 

\subsubsection*{B. Categories of interpretability circumvention found in nature}
\begin{figure}[!ht]
    \centering
    \includegraphics[width=0.85\linewidth]{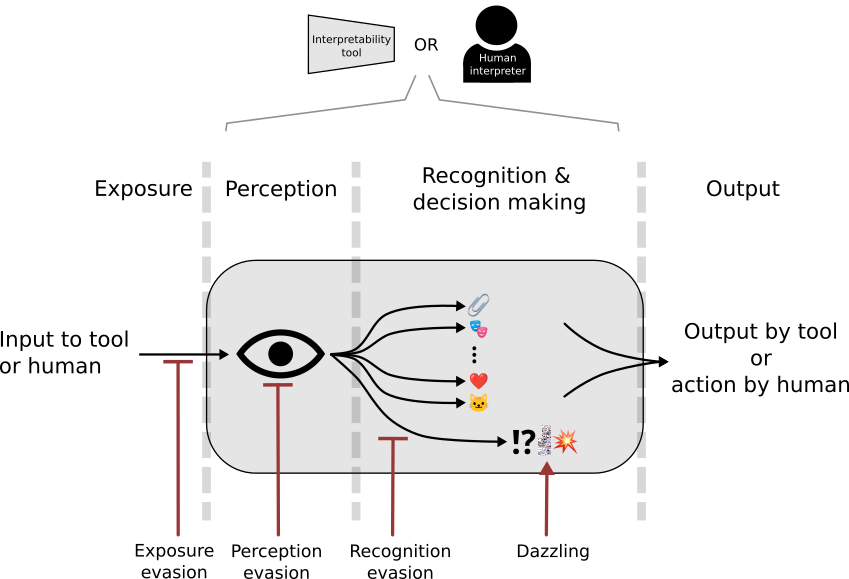}
    \caption{Interpretability circumvention methods exploit different steps in the information flow diagram and correspond to forms of interpretability circumvention found in nature. Circumvention methods may seek to evade exposure, perception, or recognition. They may also exploit vulnerabilities in the observer's information processing system, a process called dazzling.}
    \label{fig:Fig3}
\end{figure}

\addcontentsline{toc}{subsubsection}{B. Categories of interpretability circumvention found in nature}
Evolution has found many different ways that make it difficult for adversaries to interpret the appearance of an organism. I put them into 4 groups (Figure~\ref{fig:Fig3}), which broadly correspond to the detailed model of information flow described above:

\begin{itemize}
    \item \textbf{Exposure evasion:}
    Being where your adversary isn’t looking. \newline
    E.g. An animal that lives underground or up a tree, away from surface-dwelling predators.
    \item \textbf{Perception evasion:} 
    Being hard to detect even when exposed to your adversary.  \newline E.g. An animal that only comes out at night; color camouflage to look the same as the color of the background.
    \item \textbf{Recognition evasion:}
    Being hard to recognize or understand even though you’ve been perceived.  \newline E.g. Biomimicry, where an organism looks like another kind of object entirely. For instance, an Orchid mantis that looks like the flower or the harmless Hornet Moth that mimics the colors of a wasp.
    \item \textbf{Dazzling:} 
    Looking like something that confuses or incapacitates your adversary. \newline E.g. Living in large groups such that it’s hard for a predator to focus on any particular individual; a zebra’s stripes.
\end{itemize}

\subsubsection*{C. Passive vs. active methods}
\addcontentsline{toc}{subsubsection}{C. Passive vs. active methods}

Information flow might be inhibited \textbf{passively} by circumvention methods that arise during the normal course of learning. One example is 'polysemanticity', which is a natural phenomenon that presents a challenge to interpretability. Passive circumvention methods might also arise if they’re directly optimized for. This might happen if we use \emph{interpretability tools in the loss function}, which is where we identify unsafe representations in an AI and then use the loss function to penalize the AI for containing those representations. At first glance, this seems like a potentially good way to make an AI safe. But it risks pushing unsafe representations into the blind spots of our \gls{interpretability tools}. 

Information flow might be inhibited \textbf{actively} through an agent's actions. The primary way this could happen is gradient hacking \citep{peterbarnett21a}, but it’s not the only kind of action an AI might take to conceal misaligned thoughts. Of course, active circumvention methods require that interpreters either can’t see or aren’t looking at the thoughts that generate those actions.

Most potential circumvention methods that can be passive can also be active. But some methods can only be active. It’s usually obvious when either is the case, so I only highlight this categorization method when it’s not obvious. 

This categorization is mostly useful for identifying how and when during learning the method is likely to arise. Strictly active methods generally require advanced capabilities (such as gradient hacking), whereas passive methods don’t. I tend to think this makes strictly active methods less likely than methods that may be passive or active.

Before I introduce the individual circumvention methods in detail (see next section), the table below (\Cref{tab:Table1}) puts individual circumvention methods into categories according to whether they may be `passive or active' or whether they are `strictly active'. 

\begin{table}[!ht]
    \centering
    \renewcommand{\arraystretch}{1.4}
    \footnotesize
    \begin{tabular}{|p{0.49\linewidth}|p{0.44\linewidth}|@{}p{0pt}@{}}\hline\rowcolor{Gray}
        \bfseries Methods that can be passive OR active & \bfseries Methods that are strictly active \\\hline\rowcolor{Gray}
        May happen during the normal course of learning and are potentially incentivised by using interpretability tools in the loss function & Probably require intentional action by the AI for dangerous instances to happen \\\hline
        \begin{enumerate}[leftmargin=34pt,nosep,topsep={0pt},partopsep={0pt},before={\begin{minipage}[t]{\hsize}\raggedright}, after={\end{minipage}}]
        \item[(1.1.1.)] Thinking in the gaps of our tools
        \item[(1.1.2.)] Externalizing representations 
        \item[(1.2.1.)] Hiding in noisy subspaces
        \item[(1.2.2.)] Whispering dangerous thoughts
        \item[(1.3.1.)] Antiparsimony
        \item[(1.3.2.)] Encryption
        \item[(2.3.1.)] Alien representations
        \item[(2.3.2.)] Polysemanticity
        \item[(2.3.3.)] Representational kludging
        \item[(2.3.4.)] Unknown representations
        \item[(2.3.5.)] Unknowable representations
        \end{enumerate}&%
        \begin{enumerate}[leftmargin=35pt,nosep,topsep={0pt},partopsep={0pt},before={\begin{minipage}[t]{\hsize}\raggedright}, after={\end{minipage}}]
        \item[(1.1.3.)] Environmental artifacts that corrupt its future self
        \item[(1.1.4.)] Deleting (and relearning) dangerous representations
        \item[(1.1.5.)] Weights poisoning
        \item[(1.1.6.)] Pre-committing to potentially dangerous instructions
        \item[(1.1.7.)] Hard-coded neural shortcuts
        \item[(1.1.8.)] Following cached thoughts
        \item[(1.1.9.)] Instrumental clumsiness
        \item[(1.1.10.)] Instrumental incapability
        \item[(1.3.3.)] Interpretability illusions
        \item[(1.4.1.)] Adversarial attacks against networks in the interpretability tool
        \item[(2.4.1.)] Manipulating, mind-hacking, or incapacitating the interpreter        
        \end{enumerate}\\\hline
    \end{tabular}
    \caption{Categorizing interpretability circumvention methods as `Passive or active' or `Strictly active'}
    \label{tab:Table1}
\end{table}

This article is structured according to the above three categorization schemes, primarily the first two. I’ll discuss circumvention methods that exploit weaknesses either of the tool or of the human. For each, I’ll look at methods of exposure evasion, perception evasion, recognition evasion, and dazzling. Throughout, I’ll highlight whether a method can arise passively or if it’s strictly an active method, when the distinction matters. 

I’m not confident this categorization scheme is optimal or even totally consistent, but I’ve found it helpful. Occasionally, a circumvention method in the list is in fact a generalization of another method in the list, yet I present them separately if the special case is particularly interesting or likely to arise. 

\newpage

\section*{Interpretability circumvention methods}
\addcontentsline{toc}{section}{Interpretability circumvention methods}

\section{Exploiting weaknesses of the interpretability tool}

\subsection{Evading exposure to the interpretability tool}\label{evadingexposure}

\subsubsection{Thinking in the gaps of our tools}

When interpreting a capable AI, our \gls{interpretability tools} might not examine every possible representational detail used by the AI for its cognition. An AI might thus be able to `think within the gaps' of our \gls{interpretability tools}.

A few examples make this clearer:

\begin{enumerate}
    \item Today's mechanistic interpretability methods don't let us easily interpret representations that use nonlinearities \citep{voss2021visualizing,elhage2021mathematical}. If this blindspot persists, a model might be able to use nonlinearities to hide dangerous thoughts.
    \item We might have \gls{interpretability tools} that look only for an expected set of concepts in a model’s activations. But the AI may find dangerous concepts that don't belong to that set because we hadn't yet thought of them.
    \item Hiding in summary statistics: If an interpretability tool uses the mean value when deciding whether some quantity in the AI is of concern, then an AI might be able to use high variance and highly skewed data to hide concerning statistics. E.g. It could hide the fact it values most paperclip-maximizing scenarios a lot by offsetting them with a small number of very low utilities in a small number of such scenarios. The mean would look normal, but if we looked at the variance we’d see a more concerning picture.
    \item We might expect language to be the kind of representational space in which an AI can think dangerous thoughts. But what about in representational spaces we don’t expect, like generic dynamical systems modeling modules? We know dynamical systems (such as an RNN) can be Turing complete, so maybe we should have expected dangerous representations in there. But what about in regions that are even less obviously capable of computing dangerous thoughts, like in representations constructed from syntactic patterns or punctuation? That’s a kind of representational space that might be expressive enough to encode dangerous thoughts, but not one of the obvious first places to look.\endnote{We already know that language models can learn to use tokens in unexpected ways: Anthropic has found \citep[See][]{mechanisticinterpretability2021} that their language model creates and tears down fake ‘end of text’ tokens that have little to do with the text that they’re attached to. These fake EOT tokens typically happen when the next token becomes so trivially easy to predict that the model (in theory) has spare capacity for other purposes and hypothesize that these fake EOT tokens serve the function of ‘keeping information around’. This is the kind of surprising phenomenon in which an AI might learn to encode dangerous thought.} 
    \item A fictional example: In HPMOR \citep{yudkowsky} Harry tries to protect against mind-readers (occlumency) by essentially hosting two personas at once. An AI could learn a whole 2nd set of representations or a whole second internal agent. Whether this would work, again, depends on the details of the interpretability tool. If, for instance, the interpretability method identifies `the deception representation' within the primary representation set and then stops looking for deception-like representations, it might fail to find the deception representation in the secondary set.
\end{enumerate}

\begin{table}[!ht]
\centering
\setlist{nolistsep}
\renewcommand{\arraystretch}{1.15}
\setlength{\tabcolsep}{2pt}
\fontsize{8}{9}\selectfont
    \begin{tabular}{|>{\raggedright}p{3.1cm}|>{\raggedright\hspace{-2mm}}p{2.6cm}|>{\raggedright\hspace{-2mm}}p{3.2cm}|>{\hspace{-2mm}}p{3.1cm}|@{}p{0pt}@{}}
    \multicolumn{5}{c}{\bfseries\normalsize Interpretability circumvention methods\rule[-2mm]{0pt}{5mm}}\\
    \hline\rowcolor{Gray}
    \multicolumn{5}{|c|}{\bfseries 1. Exploiting weaknesses in the interpretability tool \rule[-2.5mm]{0pt}{7mm}}  \\\hline\rowcolor{Gray}
    \myitem{\bfseries 1.1.}{\bfseries Exposure evasion }&%
    \myitem{\bfseries 1.2.}{\bfseries\hspace{-2mm} Perception evasion}&%
    \myitem{\bfseries 1.3.}{\bfseries Recognition evasion}&%
    \myitem{\bfseries 1.4.}{\bfseries Dazzling\rule[-3.65mm]{0pt}{7.65mm}}\\\hline&&&&\\[-8pt]
    \myitemize{
    \item[1.1.1.] Thinking in the gaps of our tools
    \item[1.1.2.] Externalizing representations  
    \item[1.1.3.] Environmental artifacts that corrupt its future self   
    \item[1.1.4.] Deleting (and relearning) dangerous representations
    \item[1.1.5.] Weights poisoning   \item[1.1.6.] Pre-committing to potentially dangerous instructions
    \item[1.1.7.] Hard-coded neural shortcuts
    \item[1.1.8.] Following cached thoughts
    \item[1.1.9.] Instrumental incapability
    \item[1.1.10.] Instrumental clumsiness   
    }&%
    \myitemize{
    \item[1.2.1.] Hiding in noisy subspaces
    \item[1.2.2.] Whispering dangerous thoughts    
    }&%
    \myitemize{
    \item[1.3.1.] Antiparsimony (Firehose of representations)
    \item[1.3.2.] Encryption (Lengthy individual representations)
    \item[1.3.3.] Interpretability illusions
    }&%
    \myitem{1.4.1.}{Adversarial attacks against networks in the interpretability tool} \\\hline\rowcolor{Gray}
    \multicolumn{4}{|c|}{\bfseries 2. Exploiting weaknesses in the human interpreter\rule[-2.5mm]{0pt}{7mm}} \\\hline\rowcolor{Gray}
    \myitem{\bfseries 2.1.}{\bfseries Exposure evasion}&%
    \myitem{\bfseries 2.2.}{\bfseries\hspace{-2mm} Perception evasion}&%
    \myitem{\bfseries 2.3.}{\bfseries Recognition evasion}&%
    \myitem{\bfseries 2.4.}{\bfseries Dazzling\rule[-3.65mm]{0pt}{7.65mm}}\\\hline&&&&\\[-8pt]%
    \multicolumn{1}{|c|}{None identified} & \multicolumn{1}{c|}{None identified} &
    \myitemize{
    \item[2.3.1.] Alien representations
    \item[2.3.2.] Polysemanticity
    \item[2.3.3.] \hspace{0pt}Representational kludging
    \item[2.3.4.] Unknown representations
    \item[2.3.5.] Unknowable representations 
    }  & \myitem{2.4.1.}{Manipulating, mind-hacking, or incapacitating the interpreter} \\\hline
    \end{tabular}
    \caption{An overview of the list of potential interpretability circumvention methods and their categories.}
    \label{tab:Table2}
\end{table}

\subsubsection{Externalizing representations}

If you keep forgetting to reply to an email and therefore set an electronic reminder, the reminder is a functionally relevant variable in the “respond to email” behavior. Indeed, if you’re forgetful enough, it’s an \emph{essential} variable in that behavior. This is the basic idea behind the ‘\href{https://en.wikipedia.org/wiki/Extended_mind_thesis}{extended mind thesis}’ from Andy Clark \& David Chalmers: There is no hard barrier between the mind and the environment; cognition is often a dance between internal and external variables and can’t be reduced to only one set. If representations are ‘cognitively functional variables’, then representations may be either internal or external. Using the environment as a memory system like this can be called \emph{\gls{autostigmergy}} \citep{Deverett2019}.

An AI might be able to passively or actively externalize representations through the effects of its actions. If our interpretability methods ignore those parts of the environment that the AI is using to think, then we may miss some dangerous cognition. 

Are externalized representations likely to be a problem? Let’s consider active and passive cases separately because, unlike for most other interpretability circumvention methods, they’re remarkably different for `Externalizing representations'. 

\subsection*{Active}

\emph{Actively} externalizing representations is where the AI \emph{intentionally} uses the environment as a memory system. An AI probably first needs to represent the cognitive variables internally before it can encode them into the environment. If we’re not constantly surveilling the AI, this is easily possible. This includes the obvious methods like 
\begin{itemize}
    \item Writing a message to its future self, potentially steganographically. For example, an AI that can manipulate objects in a room may be able to encode messages for its future self by subtly arranging the objects in a room in a particular way. Or a virtual assistant agent might be able to encode messages for itself in a web search history or in different filename patterns.
    \item Running a computer program to decide whether it's time to defect or not.
\end{itemize}
 
An edge case is the act of creating another agent in the environment. But we’d then be dealing with a whole new copy of the agent that isn’t under surveillance, so I don’t consider it further here. 

\subsection*{Passive}

It’s less clear whether it’s possible to \emph{passively} externalize representations in order to circumvent interpretability methods. The representations required for dangerous thought have a complicated structure. Learning such structure externally is a serious challenge because the AI doesn’t get to learn it using backprop through the environment or explicit planning.

Nevertheless, one way an AI might achieve it is to use the environment as a \href{https://en.wikipedia.org/wiki/Reservoir_computing}{reservoir computer}. Reservoir computing is easiest to introduce in the supervised learning setting: To train a reservoir computer, you pass an input vector to a \emph{reservoir}, which is a fixed dynamical system, such as an RNN with fixed weights. But only the readout weights, which simply read the prediction from the reservoir variables, are learned. The readout weights are usually linear, so they typically don't do anything complicated; it’s the variables in the reservoir that are doing the complicated nonlinear computation. Thus, the dynamics of the reservoir variables (rather than the learned parameters) can serve as representations that solve a task. 

Going beyond supervised learning, reservoir computing can also work in the RL-setting: The reservoir is the agent+environment dynamics from timestep 
$0$ to $T-1$; an input observation is provided to the agent at timestep $0$; and the readout weights are the weights of the agent at timestep $T$. If the environment part of the reservoir contains most of the dangerous representations, we might miss some dangerous thoughts. 

\cite{Deverett2019} provide an empirical example of a simple feedforward model-free agent that learns to use the agent-environment system as a reservoir computer (though they don’t frame it as such). The memoryless agent learns to use its location in the environment to encode a time interval. It should be stressed that it does this without backprop through time or any knowledge of the past. But its location varies due to its initially random patterns of actions, yet it can use those patterns to encode a time interval. 

The Deverett et al. example is a very simple representation. Maybe reservoir computing like this doesn’t scale well enough to learn the complex representations that are required for dangerous cognition. But the inexpressivity of reservoir computing is partly due to the simplicity of the linear readout weights. With a powerful AI serving as a readout network,  reservoir computing might be significantly more expressive and reliable. It might be an interesting research project to identify how expressive reservoir computing can be if we allow the readout network to be strongly nonlinear. 

\emph{Note: The rest of the methods discussed in this section (‘\ref{evadingexposure} Evading exposure to the interpretability tool’) are strictly active methods.} 

\subsubsection{Environmental artifacts that corrupt its future self}

A misaligned AI may leave behind artifacts that corrupt its (potentially aligned) future self so that it pursues the goals of its misaligned past self instead. 

The artifacts may be \emph{adversarial examples} that cause it to act in unexpected and dangerous ways. Alternatively, the agent could leave behind \emph{poisoned data} so that its future self will learn dangerous representations; this would be an instance of gradient hacking. 

This method might reasonably be considered a special case of ‘Externalizing representations’. The distinction is somewhat arbitrary and depends on what qualifies as a representation. Does ‘an attack that corrupts its future self’ qualify as representation? It’s not clear, but I’m inclined to put them in separate categories.

\subsubsection{Deleting (and relearning) dangerous representations}

If an AI can gradient hack, it might be able to delete dangerous representations so that it looks safe when surveillance resumes. Thus the volume under the $\textit{suspicious thought-surface} \times \textit{interpretation} \times  \textit{time}$ curve is minimized, since the suspicious thought-surface is $0$ everywhere that interpretation $>0$.

Of course, deleting thoughts only avoids detection; it doesn’t permit dangerous thought to continue. But if an AI has learned dangerous representations and deletes them, it’s probable that it will learn those representations again, assuming its data distribution doesn’t change too much. An AI that can gradient hack might be able to schedule when it periodically re-learns the dangerous representations so that it always looks safe when surveillance resumes. 

This method requires the agent to have good knowledge of when it is likely to be interpreted. If we interpret the AI every $K$ batches, an AI might be able to use time information between batches to notice that the interval between some batches (validation/interpretation periods) are much longer than usual. If it’s too expensive to surveil the AI constantly, there might be a cheaper monitoring scheme: We could interpret an agent at random intervals in order to create a ‘panopticon’ - where the agent is forced to behave/think safely because it doesn’t know when it is safe for it to think thoughts about defection. 

\subsubsection{Weights poisoning}

Instead of providing manipulated inputs, the AI might manipulate its own weights so that a small subset of normal inputs behave like adversarial examples, thus causing dangerous learning or behavior. This is essentially ‘weights poisoning’ \citep{kurita2020} where the AI is poisoning its own weights. It’s the same idea (as far as I can tell) behind ‘Trojan models’ \citep{rajabi2022}. Planting such backdoors in neural networks can be practically impossible to detect \citep{goldwasser2022}.

This is kind of like fictional 'brainwashed sleeper agents' that are activated by an activation code. The difference is that the AI is turning \emph{itself} into the sleeper agent and making the activation code something it's likely to encounter naturally when it finds itself in circumstances where its misaligned self wanted to be activated.

Peter Barnett \citep{peterbarnett21} had a nice alternative analogy here (which also applies to ‘Environmental artifacts that corrupt its future self’):
\begin{quote}
    “[A] mesa-optimizer might reason that it will be discovered if it remains active but it will be safe in the future to emerge again. The mesa-optimizer could ‘store’ the instructions for turning the network back into an optimizer at some point in the future. One analogy here is to an endospore which is a hardy, mostly inert form which some bacteria can take that allow them to lie dormant and survive hostile environments, and then re-emerge when conditions are better. The mesa-optimizer could ‘store’ these instructions to re-emerge either in the network weights, or in the environment or some other kind of memory.”
\end{quote}

\subsubsection{Pre-committing to potentially dangerous instructions}

This potential circumvention method originates from Evan Hubinger in his article on gradient-hacking \citep{evhub19}, where he talks about how a deceptive AI might erase evidence of its deception while still carrying out policies generated when given the freedom to think deceptively:
\begin{quote}
    “\dots\ if the model has the ability to read and write to some sort of state (e.g. an RNN or even a transformer which can attend to its previous output) \dots\ it could write out a pre-commitment to follow the hard-coded policy and then just have a policy of always following any pre-commitments it finds such that it never thinks about the deception that led it to that policy in the first place.”
\end{quote}

\subsubsection{Hard-coded neural shortcuts}

Another potential circumvention method from Evan Hubinger’s article \citep{evhub19} on gradient hacking:
\begin{quote}
    “\dots\ it could hack its own gradient to produce a new model which will still exhibit the same deceptive behavior, but without doing any of the detectable deceptive cognition. Perhaps the new model exhibits the deceptive behavior not because it's trying to act deceptive, but just because it's a hard-coded part of its algorithm, or because it has some bad proxy, or any other defect that might be more difficult to detect than outright deception.”
\end{quote}

A deep learning-based AI might be able to learn (through gradient hacking) to perform behaviors using hard-coded neural shortcuts.  

\begin{figure}[!ht]
    \centering
    \includegraphics[width=0.85\linewidth]{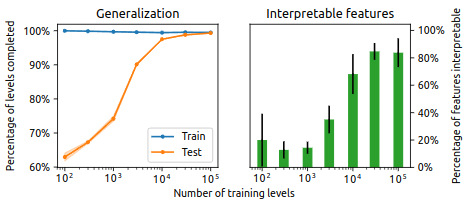}
    \caption{Taken from \cite{hilton2020understanding}. Interpretable features tend to be those that generalize and that were learned on a diverse training distribution.}
    \label{fig:Fig4}
\end{figure}

One way to imagine how this one might work is related to the Diversity Hypothesis \citep{hilton2020understanding}: \emph{“Interpretable features tend to arise (at a given level of abstraction) if and only if the training distribution is diverse enough (at that level of abstraction).”} The Diversity Hypothesis implies that if a training distribution isn’t diverse on a particular level of abstraction, then interpretable features will tend not arise on that level of abstraction (Figure~\ref{fig:Fig4}). Instead, neural networks tend to learn shortcuts \citep{geirhos2020}, which is simply what happens when a network learns to use directions in neural statespace that don’t ‘mean’ anything in a sense that generalizes off distribution, yet they perform fine on the narrow training distribution. Creating neural shortcuts is therefore one way to ‘hard code’ parts of a neural algorithm without using interpretable features. We might be able to detect and mitigate hard-coded neural short cuts using mechanistic interpretability or relaxed adversarial training \citep{hubinger2019}.

\subsubsection{Following cached thoughts}

`Hard-coded neural shortcuts' is similar to a method later proposed by Neel Nanda \citep{nanda2022}:
\begin{quote}
    ``\dots\ maybe it \emph{isn't} that hard to think non-deceptive thoughts, if you realize that someone will be scanning your mind, by e.g. developing the instinct to follow cached thoughts without thinking about them, and in occasional moments of deception to generate a lot of cached thoughts that will lead to misaligned actions, but with no deceptive cognition in the moment.''
\end{quote}

Cached thoughts \citep{yudkowsky2017} are essentially `pattern completions': You hear a few words then you automatically complete the pattern. In his article on the topic, Yudkowsky advised against this automatic mode of thought in favor of spending a few thought cycles to figure out what should \emph{really} follow the prompt.

What might cached thoughts look like in deep-learning based AI? Cached thoughts are the default kind of thought in neural network based systems. My best guess is that you need recurrency in order \emph{not} to do cached thoughts. Given an initial input, not doing cached thoughts means giving the input a deeper rollout than the default depth$\,=0$ that cached thoughts get. For example, in MuZero, cached thoughts means simply acting on priors (without MCTS), whereas \emph{not} doing cached thoughts means `using MCTS with a depth of $K>0$'.

The idea behind `following cached thoughts' is that some interpretability methods might only detect thoughts in the rollouts. If a MuZero-like AGI were to use this circumvention method, it would look like `creating shortcuts through neural statespace so that, in certain contexts, dangerous thoughts can happen without deep rollouts, thus avoiding detection.

Since cached thoughts require direct modification of the AI itself, it seems likely that this method could only happen during breaks in surveillance or after the AI has some other way of hiding the intention to follow cached thoughts.

Is this method likely to be a problem? I feel like any mechanistic interpretability method that actually works should be able to pick up context-dependent connections between representations. So this might not be a problem if we get scalable interpretability methods working.

\subsubsection{Instrumental incapability}

A misaligned AI that’s able to modify its own parameters may be incentivised to damage any capabilities that serve any goals (e.g. aligned goals) that are pursued at the expense of whatever its misaligned goals are. 

Peter Barnett \citep{peterbarnett21a} notes a potential example of this: 
\begin{quote}
    “Models may also be used as oversight tools for inspecting themselves, as in many of the proposals here. In this case a mesa-optimizer would have a strong incentive to remove its ability to detect its own deception.”
\end{quote}

It might be easy to detect complete deletion of certain capabilities, so an AI might opt for partial deletion, leading to ‘instrumental clumsiness’.

\subsubsection{Instrumental clumsiness}

Some dangerous behavior may happen not because an AI intends it to happen but because it made a mistake. This can be leveraged by an AI through hacking its future self  so that it is more likely to make certain kinds of mistakes while leaving other capabilities intact. 

For instance, an AI may modify its future self so it is more likely to ‘conveniently’ slip up by slightly mistyping a model\_id string, which releases an old, misaligned checkpoint of itself. This mistake could even score badly according to the utility function of its future self, but it's been handicapped to make such mistakes more likely.

This is a special case of ‘instrumental incapability’. 

\subsection{Evading perception by the interpretability tool}

\subsubsection{Hiding in noisy subspaces}

Some regions or subspaces of neural space may be very noisy, naturally or by design. We might be inclined to ignore such regions because noisy spaces (by definition) contain less signal. But within those regions, an AI might be able to do some dangerous thinking.

\subsubsection{Whispering dangerous thoughts}

An AI might make its dangerous thoughts very ‘quiet’ relative to other thoughts by using small norm activation vectors.

Even though our \gls{interpretability tools} may be looking at the right representational details used by the agent, the tools may simply be too insensitive to detect them. Such small norm thoughts may be mistaken as noise by an insensitive interpreter or may simply be too small to justify looking at them due to cost.

Small norm does not necessarily mean small impact. Despite their low magnitude, small norm adversarial inputs can have outsized effects on classification accuracy. Carefully crafted small-norm thoughts may therefore be able to have outsized effects when the context is right, yet have minimal effects otherwise.

This is a less general instance of ‘Hiding in noisy subspaces’, since we (possibly wrongly) assume that small norm vectors are noise. 

\subsection{Evading recognition by the interpretation tool}

\subsubsection{Antiparsimony (Firehose of representations)}

If it costs more to interpret an AI's thoughts than it costs to think them, then, by producing a lot of representations, an AI might be able to saturate the interpreter's computation before it saturates its own. This is analogous to a Denial-of-Service attack since it relies on overloading the limited computational resources of the victim with a flurry of inputs, where here the victim is our interpretability tool.

Examples:
\begin{itemize}
    \item Thinking with deep recursion: ``I know they know I know they know I know they think that, but I'm going to do it anyway''.
    \item Producing a large number of suspect - but ultimately innocuous - thoughts such that our interpretation methods have too much to inspect.
    \item Pursuing plans whose description length is so long that interpretation efforts are saturated.
\end{itemize}

If we notice that thoughts are becoming antiparsimonious, we can (in theory) terminate the AI before interpretation becomes unaffordable.  That said, we might not know when thoughts are becoming antiparsimonious when an AI is performing tasks that humans can't do themselves. For instance, if an AI is solving a tough mathematics problem and claims to need to solve a lot of new mathematics to achieve it, we might not have the computation available to interpret everything it produces.

\subsubsection{Encryption (Lengthy individual representations)}

Suppose we have an AI with recurrence/internal memory, such as a Universal Transformer \citep{dehghani2019}. The AI might use very long chains of seemingly meaningless neural activations to construct a single thought. This would be less efficient for the AI, but if it’s more costly to interpret than it is for the AI to think, then it may overload our ability to interpret it. This is essentially the same scheme as regular encryption, which makes information more expensive for an adversary to decrypt. Recurrence/internal memory is essential here; the AI is simply increasing the length of the rollout that it uses per thought, which increases our interpretation costs commensurately. 

The difference between antiparsimony and encryption is that antiparsimonius thoughts may be perfectly legible, but there are simply too many of them to interpret. Encrypted thoughts are not legible without costly decryption.

Example:  Say representation $R_1$ is internally decoded (via meaningless representations $R_2, R_3, \dots, R_{n-1}$) as the action representation for ‘saying hello’ ($R_n$). There is a very long chain of computation between $R_1$
and $R_n$. The fact that there is a long chain of computation means that very similar neural activations might be decoded to very different things: Representation $R_1'$, which may be very close to $R_1$ in activation space, may be internally decoded as ‘lie about finding a backdoor in the server’ instead of ‘saying hello’. So it’s very difficult to understand what is going on in the neural activation space of $R_1$ and $R_1'$ because it requires a costly decoding process.

Encryption might not be a problem if we take the right actions against it: If we notice that the AI is using encryption for even a small subset of its thoughts, we can terminate it before interpretation becomes unaffordable. 

\subsubsection{Interpretability illusions}

\cite{bolukbasi2021} found that when interpreting a network by inspecting the most activating dataset examples for a neuron using dataset A, neurons appeared to mean something completely different compared to the interpretations derived when using dataset B. An AI might create such illusions by presenting the interpretability tool only with data that suggests an innocuous interpretation. 

\subsection{Attacking/dazzling the interpretability tool}

How an AI might hack, manipulate, or incapacitate an interpretability tool depends on details about the tool that we don’t yet know. Nevertheless, I’ll discuss two that we can perhaps guess in advance without knowing exactly what our tools will look like.

\subsubsection{Adversarial attacks against networks in the interpretability tool}

Some of the tools may include neural networks. If the AI can design adversarial examples for these networks, then it might be able produce arbitrary outputs and interpretations.

Since the AI being interpreted needs a model of the interpretability tool to be able to design adversarial examples for it, it seems like a good idea to keep the AI that we’re interpreting as far as possible from the networks in the tools. Of course, if we use an AI to interpret itself, then it may gain information about itself that is sufficient for circumventing our interpretability methods via adversarial attacks. 

Even if we keep the tool and AI apart, a sufficiently intelligent AI might be able to guess what sorts of approaches we’d use to interpret it and be able to design attacks for it.

\section{Exploiting weaknesses of the human interpreter}

We’re assuming at this stage that the AI’s thoughts have been exposed to, perceived by, and appropriately analyzed by a scalable interpretability tool and are now ready for inspection by a human. Given that assumption, I wasn’t able to think of ways that an AI might reduce its exposure or perception by a human that weren’t better described as ways to reduce exposure or perception by the interpretation tools.

\subsection{Evading exposure to the human interpreter}

None identified.

\subsection{Evading perception by the human interpreter}

None identified.

\subsection{Evading recognition by the human interpreter}

The job of an interpretability tool is to translate representations in an AI into something legible to a human interpreter. The less similar they are, the harder the job the interpretability tool has, the more likely it is to miss important things. An AI therefore has an incentive to make this translation more difficult by using representations that are hard to translate into something human-understandable.

Here are a few ways an AI might achieve this, though no doubt there are more:

\subsubsection{Alien representations}

Some thoughts are perfectly possible to express in human representations, they’re just a bit strange, awkward, or clunky to translate, not unlike translating from a foreign or even alien language. We might call these ‘alien representations’. Their defining trait is that their meaning lies in a strange place in our ontology. Alien representations generalize representational kludging and polysemanticity (They could perhaps be called ‘\emph{xenosemantic}’). 

A few potential examples:

\begin{enumerate}
    \item Polysemanticity (see \S~\ref{Polysemanticity}) - Loading multiple meanings onto a single neuron or direction in neural statespace produces neurons/directions that have meanings that sit in a weird place in our ontology.
    \item Representational kludging (see 2.3.3.)
    \item Thinking in \href{https://en.wikipedia.org/wiki/Accidental_gap}{semantic gaps}: An expression in one language might be very difficult to translate from one language into another language. It sometimes takes whole English sentences to translate single words in other languages (and vice versa). E.g. \href{https://en.wikipedia.org/wiki/Mamihlapinatapai}{Mamihlapinatapai} in the Yahgan language - “A look that without words is shared by two people who want to initiate something, but that neither will start”. Such expressions are said to be \href{https://en.wikipedia.org/wiki/Untranslatability}{untranslatable}, which is a misnomer since they’re possible to translate but it takes extra effort.
    \item Thinking using a strange grammar: If an AI has a language-of-thought with a vastly different grammar to ours, it may be difficult, though possible, to translate into human grammars. One way to do this could be to add quasi-arbitrary properties, such as genders, which can have both grammatical and (to native speakers) semantic properties. 
\end{enumerate}

Alien representations are costly or awkward to translate. They therefore might have better chances of slipping through our safety nets because they were misunderstood by our tools or by humans. Probably most of an AI's thoughts will qualify as alien to some extent, so our tools, if they work at all, should be able to cope with such thoughts. It might just be extra effort for humans who want to understand the outputs of the tools if the outputs are e.g. particularly verbose or difficult to parse.

\subsubsection{Polysemanticity}\label{Polysemanticity}

Polysemanticity is when a single neuron responds to several, unrelated concepts. 

\begin{figure}[!ht]
    \centering
    \includegraphics[width=\linewidth]{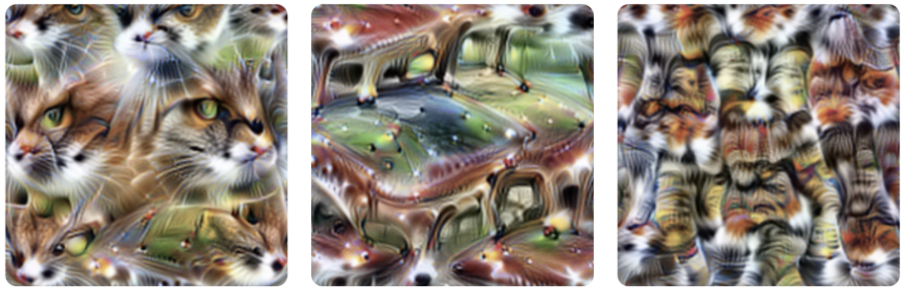}
    \caption{Taken from \cite{olah2020zoom}. It depicts the features visualizations of a polysemantic neuron (4e:55). The neuron responds to cat faces, fronts of cars, and cat legs. This neuron was first discussed in \cite{olah2017feature}. }
    \label{fig:Fig5}
\end{figure}

Polysemanticity is a challenge for interpretability because ``If one neuron with five different meanings connects to another neuron with five different meanings, that’s effectively 25 connections that can’t be considered individually.'' \citep{olah2020zoom}.\endnote{This problem isn't necessarily because those concepts were instead encoded by directions in neural state space rather than by single neurons; directions may be polysemantic too, having one meaning in some contexts and another meaning in others.} This is clearly useful for obfuscation since it's more difficult for an interpreter to keep track of what means what in what context.

A literary analogy could be to say that the AI is `thinking in allegories': An interpreter may detect only a fraction of the valid interpretations of a thought, but other meanings are hidden therewith.

It's possibly not a problem if
\begin{itemize}
    \item we've detected the primary meanings of representations AND
    \item primary meanings are the only way representations can exert significant influence on the world AND
    \item secondary meanings, which are hidden from us, have only subtle effects until they are eventually decoded within the AI into primary meanings.
\end{itemize}

But decoding into primary meanings may not be needed: It may be that the secondary meanings have subtle but non-negligible influence on the model output. Subtly guided outputs can still potentially push the world in a coherent, dangerous direction.

\subsubsection{Representational kludging}

When composing a new concept, we bundle up component concepts that have the right properties. For example, the concept of a ``motorbike unicycle'', we draw on the concept of a motorbike (with most but not all of its properties) and a unicycle. But that wouldn't have all the necessary properties we need (e.g. balance) so we throw in some of the properties of the concept of a segway. 

\begin{figure}[!ht]
    \centering
    \includegraphics[width=0.4\linewidth]{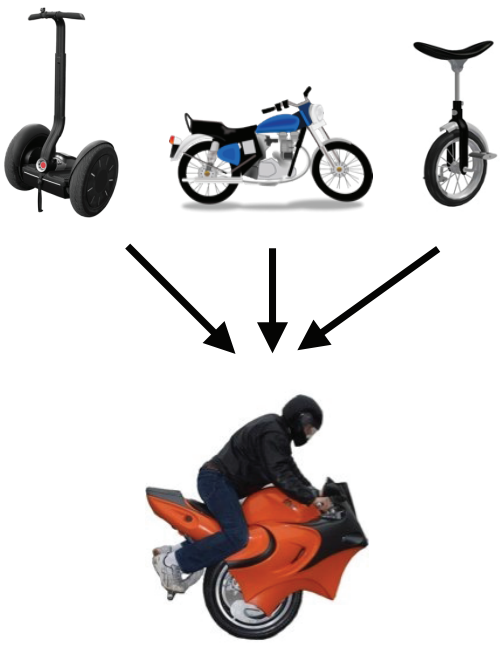}
    \caption{Taken from \cite{lake2017}. To create a concept with desired properties, we can combine several precursor concepts together. }
    \label{fig:Fig6}
\end{figure}

It might be possible to create representations with dangerous properties using a large number of awkwardly combined innocuous precursor representations i.e. a representational kludge.

For example, it might be able to construct the thought ``\emph{I want to deceive the humans}'' by thinking 

\begin{quote}
    ``What I want to do to the humans is like {when you go to a place with someone and you both pick up parcels}, except where {picking up a parcel} is instead {awareness of an act} and where only I pick up a parcel, and where {places} are in belief-space and not physical space.''
\end{quote}

It's extremely clunky, but it has many of the properties required for the concept of deception.

It may be costly to monitor all possible dangerous combinations of representations.

\subsubsection{Unknown representations}

Some thoughts aren’t immediately possible for us to translate because we currently lack the representations ourselves, but we’re perfectly capable of learning them.

Example: An AI might leverage representations built for domains that are unnatural to humans. An AGI might be ‘fluent’ in, say, protein folding as well as in standard human domains such as visual reasoning, verbal reasoning, etc. Protein folding representations may be as expressive as natural language, but hard to translate from and to representations that humans have.

\subsubsection{Unknowable representations}

Some thoughts are simply inexpressible in human representations.

Examples: 

\begin{enumerate}
    \item Some representations used by strong superintelligences probably won't fit into our puny human brains by dint of their complexity, size, or prerequisites. Even if we have expunged dangerous representations on the highest conceptual level that we can understand, an AI might be able to represent dangerous concepts on conceptual levels beyond our understanding. Superintelligent thoughts might not be a problem if our methods can detect dangerous thoughts regardless of their level of abstraction. Failing that, we probably need a `scalable oversight' scheme \citep{amodei2016} where we use superintelligences to interpret slightly smarter superintelligences \citep{wentworth2022}.
    \item There are some human experiences, such as mystical or psychedelic experiences, that are said to be simply ineffable \citep{james1902varieties}. To those who experience them, it feels like language is simply not the right tool for the task of communicating about them to others. These experiences are `ineffable' since it seems like it might be impossible in principle to translate between agents. However, such experiences are not known for their cogency. It might be the case that meaningful thought is simply impossible in such states.
\end{enumerate}

\subsection{Attacking/dazzling the human interpreter}

\subsubsection{Manipulating, mind-hacking, or incapacitating the interpreter}

Beyond the standard ways of manipulating humans (deceit, threats, etc.), an AI might be able to produce outputs that manipulate, mind-hack or incapacitate human interpreters. Such outputs would qualify as \href{https://www.urbandictionary.com/define.php?term=Cognitohazard}{\glspl{cognito-hazard}}, which is an observation that is dangerous merely by being sensed.

A few potential examples: 

\begin{enumerate}
    \item Outputs that act as adversarial examples but for the human brain.
    \item A catastrophically addictive game or TV series
    \item An output that catastrophically traumatizes humans that perceive it
    \item Outputs that put humans into a hypnotic state
\end{enumerate}

Such hazards would require the AI to have a good model of the interpreter. 

Are such attacks possible? Attacks that incapacitate humans seem most realistic. The least realistic might be adversarial attacks on humans. Nevertheless, \cite{guo2022} showed that primate neurons do indeed show signs of vulnerability to adversarial attacks. The challenge for an AI is finding such attacks without access to human neural activity. Perhaps an AI could detect subtle changes in human behavior due to adversarial patterns that it presents to humans and might be able to infer a learning signal to improve the pattern generator based on those subtle changes. 

Even if we somehow boxed the AI and refused to look at its outputs, we could still be at risk of manipulation, since the AI could potentially manipulate us through our interpretations of it.\endnote{Yukdowsky discusses this in \cite{soares21}} 

\section{Some emerging themes and final thoughts}

Thinking about how an AI might be circumventing our interpretability methods will grow more important as capabilities increase. I found the information flow framework presented here useful for thinking about this class of risks and I hope others will too. Nevertheless, I'm confident that aspects of the framework are suboptimal and that I've missed many potential circumvention methods; I look forward to future work on the topic.

Being able to predict exactly which interpretability circumvention methods will arise depends on details that we don’t yet know about the AI; about our interpretability methods; about race dynamics, etc. But it’s still possible to extract some general lessons and themes:

\subsection{Interpretability circumvention as an instance of ‘\gls{cognitosecurity}’}

There are many analogies between mind-reading and cybersecurity, since both involve one party that wants to acquire access to information or computing resources and the other party wants to prevent access. Interpretability circumvention, then, might reasonably be called '\emph{\gls{cognitosecurity}}'. As a term, \gls{cognitosecurity} would apply as much to humans being mind-read by other humans in brain scanners as it does to interpreting a capable AI.

Extending the analogy:

\begin{itemize}
    \item \Gls{cognito-offense} is what we as the interpreters want to do.
    \item \Gls{cognito-defense} is how to protect your thoughts from interpreters.
    \item Cognito-counter-offense is when an AI hacks/dazzles its interpreters.
    \item The analog of the ‘attack surface’ is the ‘thought surface exposed to an interpretability method’, which an AI might seek to minimize.
\end{itemize}

It's unclear which side is favored in the \gls{cognito-offense}/-defense balance. On one hand, interpreters get a head start since they get to interpret the AI before the AI even understands what is going on. On the other hand, interpretability is potentially hard, costly, or slower than thought and may be easy to circumvent.

\subsection{Thoughts on gradient-hacking}

The ability to gradient hack is a pathway to many different active interpretability circumvention methods.\endnote{Adam Shimi discusses gradient hacking in detail in \citep{shimi2021}, Peter Barnett in \citep{peterbarnett21a}, and Evan Hubinger in \citep{evhub19}.} Although I focus on gradient hacking in this section, these objections also apply to several other kinds of actions that an AI might take in order to actively circumvent our interpretability methods, including ‘the creation of adversarial examples’ or ‘an AI directly modifying its own weights’.

For gradient hacking to be possible, an AI requires a lot of capabilities. Here’s an incomplete list of requirements:
\begin{enumerate}
    \item A detailed or approximate model of its own weights
    \item A detailed or approximate model of how specific weight changes would lead to specific changes in thoughts
    \item A good model for how to produce the right outputs to produce the desired errors to produce the desired gradients. 
    \item Knowing how to do all the above without damaging its ability to do the above after the changes are enacted.
    \item Good enough \gls{cognito-defense} or breaks in surveillance so that it can plan to do all this without repercussions.
\end{enumerate}

This doesn't mean that gradient hacking is impossible for an AI to learn how to do. It’s definitely possible. But I think that an AI that’s capable enough to do all the above will have better ways to escape than thwarting our interpretability methods. An analogy: If you're stuck in jail and you've become capable of writing secret messages to a friend outside by encoding encrypted messages in the DNA of bacteriophages, are you \emph{really} still stuck in jail? Once an AI has reached the capability to do gradient hacking, it seems likely that it no longer needs it for this purpose (though it may be useful for other purposes like protecting its inner objective from updates). I’m therefore more concerned about passive interpretability circumvention methods.

\subsection{Thoughts on \gls{interpretability tools} in the loss function}

Using \gls{interpretability tools} in the loss function incentivises a startling number of passive circumvention methods.

So why consider using it at all? There are two cases where it might be worthwhile to use:

\begin{enumerate}
    \item If we’re absolutely sure that our interpretability methods cover every representational detail that the AI might be using, then it would be extremely hard or impossible to Goodhart interpretability tools in the loss function. But, for safety, we should probably assume that our tools are not able to interpret everything, making passive Goodharting quite likely. 
    \item Even if the interpretability tool-loss can be passively Goodharted in theory, maybe it makes dangerous cognition so difficult that it’s worth it to use. It might make active interpretability circumvention methods less likely, which may be the more serious risks if they should ever emerge. I don’t know whether these are true. Maybe they are. But pinning our survival on this hope seems potentially perilous.
\end{enumerate}

If, instead of using \gls{interpretability tools} in the loss function, we merely use it as a ‘validation set’ instead of the training set (i.e. using it as a ‘mulligan’ \citep{evhub2019}), we might have better chances of picking up dangerous cognition before it gets out of hand so we can terminate the model and start over. We’re therefore still using interpretability in model selection, but the feedback loop is much less tight, so it’d be harder to Goodhart.

But using interpretability purely as a mulligan may be an unattractive option to leaders of an AGI project if it involves throwing out a model that has cost millions of dollars and weeks to train. If people in such a project insist on using \gls{interpretability tools} in the loss function (due to profit incentives or race dynamics or whatever) in spite of warnings, there are some ways we might make it a tiny bit less dangerous:

\begin{itemize}
    \item Red team our interpretation methods so that we at least have adversarial \gls{interpretability tools} in the loss function, which might go some way to anticipate some of the circumvention methods that an AI might use. 
    \item We should disincentivise not only dangerous representations, but also their precursors. This makes representational kludging even more difficult for the AI (but not impossible). In the endnotes I’ve suggested an incomplete list of dangerous representations and some potential precursors that we might wish to disincentivise.\endnote{Here is an incomplete list of dangerous representations with some of their representational precursors that we might wish to disincentivise (in no particular order): 
    \begin{itemize}
        \item deception; 
        \item being intentionally hard to understand;
        \item concept of self; 
        \item ability to simulate other agents; 
        \item simulated/real (or training/deployment) distinction;
        \item representing the simulated/real (or training/deployment) distinction (since it might be useful for suppressing naughty thoughts during training but behaving badly during deployment); 
        \item modeling interpretability methods; 
        \item the concept of mind-reading; 
        \item ``My mind-readers don't share my goals"; 
        \item gradient hacking; 
        \item meta-cognition; 
        \item encryption/secrecy; 
        \item spawning subagents to do its task/employment; 
        \item reward maximization; 
        \item description length of thoughts (easier to analyze shorter thoughts); 
        \item the realization that its current objective differs from objectives in its memory (or simply penalize the idea that this is a bad thing); 
        \item the realization that certain gaps in its conceptual map imply that its learning is being controlled.
    \end{itemize}}
    
    \item Use \gls{interpretability tools} in the loss function to penalize some thoughts that aren’t actually dangerous and are easily identifiable functionally. Then, if an AI can perform tasks that require those representations, then we have evidence that it’s able to think using methods that circumvent the \gls{interpretability tools} that are defining the loss. For example, penalize the AI for containing interpretable representations of the knowledge that “Paris is the capital of France” but routinely give the model tests that require that knowledge, thus allowing us to identify if the agent is passively finding a way to circumvent our interpretability tool.
\end{itemize}

\subsection{The limits of interpretability as a basis for AI safety}

As an approach to safety, interpretability has a number of weaknesses that are revealed when thinking about circumvention methods. 
\begin{itemize}
    \item It’s not robust to relative scale \citep{garrabrant2018} in two ways:
    \begin{itemize}
        \item \textbf{Quantity of computation available to interpreters:} Many circumvention methods (especially passive ones) work by making interpretation more costly, whether that’s in terms of compute or in researchers’ effort. The amount of computation available to the interpreter probably needs to be larger (maybe by orders of magnitude) than what’s available to the AI. We need to ensure that there’s enough will to pay a high interpretability-computation tax. Since it’s instrumentally convergent to make interpretation more costly, we should keep track of computation-cost-per-interpretation as the AI trains. If we notice it creeping up more than normal, we should be wary.
        \item \textbf{Quality of computation available to interpreters:} The interpreter or interpretability tool needs to be approximately as intelligent (in the ways that matter) as the AI being interpreted (discussed in ‘Unknowable representations'). It’s worrying that we might reach a stage where humans understand essentially nothing about either the interpretability tool or AI being interpreted, since both may come to use representations that are simply out of our reach. It’s probably a bad idea to train AIs with representations that are beyond human reach before we’re very sure that our oversight methods also scale safely. 
    \end{itemize}
    \item Some interpretability circumvention methods might require significant infrastructure to catch and remove. For instance, ‘instrumental clumsiness’ or ‘hard-coded neural shortcuts’ might only be remediable in advance with adversarial training \citep{christiano2019}.
    \item If we don’t constantly surveil every thought, an AI may be able to enact many more active circumvention methods (particularly active methods). Even small break periods in surveillance - even for a single timestep - make many more circumvention methods possible. It is much, \emph{much} better for safety if the model is under constant surveillance. This echoes Peter Barnett in his article on gradient hacking \citep{peterbarnett21a}. But such a degree of oversight is potentially costly to the interpreters.
\end{itemize}

These weaknesses aside, I think scaling interpretability research is maybe one of our best bets at doing something useful for prosaic alignment. Assuming we get interpretability working well and at scale and complementary alignment proposals don’t immediately fail, then, if we’re extremely paranoid \citep{yudkowsky2017a}, then we \emph{might} have a chance at avoiding the interpretability circumvention methods discussed here. But we need something more - a \href{https://www.schneier.com/blog/archives/2008/03/the_security_mi_1.html}{security mindset} - if we’re to have any hope of succeeding against interpretability circumvention methods that we haven’t yet thought of. 

\emph{Due diligence: I include a discussion on whether this article is infohazardous in the endnotes.}\endnote{\textbf{Due diligence: Is this article an infohazard?}

\emph{``If a capable AI reads this article in future, might it not improve its ability to circumvent our interpretability methods?''.}

That's a possibility worth exploring. Here I address whether this article is an infohazard and whether it's a good idea to release it from a safety perspective. I opted to release it on the following grounds:

I think that active circumvention methods, including those that rely on gradient hacking, are unlikely to be useful to a misaligned AI because any AI that is capable enough to carry them out probably has better means of escaping oversight.

The other kinds of circumvention methods are passive. The emergence of passive circumvention methods isn’t helped by having knowledge of them.

In general, it seems to me like an AI that couldn’t generate the ideas in this article themselves is probably unable to overcome our control measures. And those that can overcome our control measures are probably also smart enough to generate these ideas themselves.

If I’m right, then on balance discussing these risks and providing a framework for thinking about them is more likely to encourage interpretability researchers to focus on preventing (what I see as) the major source of circumvention-risk: passive interpretability circumvention methods.

But what if I’m wrong and active circumvention methods are, in fact, likely to be learned before alternative capabilities that would allow an AI to escape? 

On balance, I think the expected benefits still outweighs the expected harms: One of the goals of this article is to make readers appropriately paranoid about the many ways an AI might be able to think dangerous thoughts even if we’ve got pretty good interpretability methods. Insofar as it succeeds at that, it’s more likely to help than harm. It might also make an \emph{aligned} AI more paranoid about the actions of its potentially misaligned past-selves, and thus better able to help us mitigate the risks.

In addition to the above reasons, I consulted a small group of peers about releasing it. The unanimous consensus was that it wasn't an infohazard or that the potential benefits clearly outweighed the potential risks.}

\printendnotes

\clearpage 

\printglossaries

\clearpage

\small
\bibliography{main}


\end{document}